\documentclass{article}
\usepackage{ijcai23}

\usepackage{times}
\usepackage{soul}
\usepackage{url}
\usepackage[hidelinks]{hyperref}
\usepackage[utf8]{inputenc}
\usepackage[small]{caption}
\usepackage{graphicx}
\usepackage{amsmath}
\usepackage{amsthm}
\usepackage{booktabs}
\usepackage[switch]{lineno}

\usepackage{amsmath}  
\usepackage{fancyref} 
\usepackage[ruled,linesnumbered]{algorithm2e}
\usepackage{setspace}
\usepackage{hyperref}
\hypersetup{hypertex=true,
colorlinks=true,
linkcolor=blue,
anchorcolor=blue,
citecolor=blue}
\usepackage{arydshln}
\usepackage{multirow}
\usepackage{caption}
\usepackage{float}
\usepackage{subcaption}
\graphicspath{{Figures/}}

\title{Determinate Node Selection for Semi-supervised Classification Oriented Graph Convolutional Networks}

\author{
Yao Xiao$^a$
\and
Ji Xu$^{a}$\footnote{: corresponding author. Email: jixu@gzu.edu.cn.}\and
Jing Yang$^a$ \And
Shaobo Li$^a$
\affiliations
$^a$ State Key Labortory of Public Big Data, Guizhou
University
}

\begin{document}

\maketitle

\begin{abstract}
    Graph Convolutional Networks (GCNs) have been proved successful in the field of semi-supervised node classification by extracting structural information from graph data. However, the random selection of labeled nodes used by GCNs may lead to unstable generalization performance of GCNs. In this paper, we propose an efficient method for the deterministic selection of labeled nodes: the Determinate Node Selection (DNS) algorithm. The DNS algorithm identifies two categories of representative nodes in the graph: typical nodes and divergent nodes. These labeled nodes are selected by exploring the structure of the graph and determining the ability of the nodes to represent the distribution of data within the graph. The DNS algorithm can be applied quite simply on a wide range of semi-supervised graph neural network models for node classification tasks. Through extensive experimentation, we have demonstrated that the incorporation of the DNS algorithm leads to a remarkable improvement in the average accuracy of the model and a significant decrease in the standard deviation, as compared to the original method.
\end{abstract}

\section{Introduction}

Since graphs are effective in representating real-world objects with interactive relationships and interdependencies, recently, many studies on extending deep learning approaches for graph data have emerged \cite{wu2020comprehensive}. Graph neural networks (GNNs), which are capable of processing graph-structured data, have gained significant attention and have been widely applied in the modeling and solution of various problems across multiple fields,  including social networks \cite{kipf2016semi,hamilton2017inductive}, recommender systems  \cite{yu2021self}, knowledge graphs \cite{fang2022molecular}, and graph matching \cite{wang2020combinatorial,wang2021neural}.

Graph neural networks (GNNs) can be utilized for a variety of graph analytics tasks, including node classification \cite{kipf2016semi}, graph classification \cite{zhang2018end}, graph generation \cite{simonovsky2018graphvae}, and link prediction \cite{zhang2018link}. In the context of node classification, classical methods such as Graph Convolutional Networks (GCNs) \cite{kipf2016semi}, GraphSAGE \cite{hamilton2017inductive}, and Graph attention network (GAT) \cite{velivckovic2018graph} infer the labels of unlabeled nodes in the graph by using a small number of labeled nodes for semi-supervised learning through the feature and structural information of the graph. GCNs, in particular, have been shown to be effective in extracting structural information from graph through the use of graph convolution operations to learn node embeddings for semi-supervised classification.

However, GCNs still have some limitations. One such limitation is the issue of over-smoothing, which can occur when too many GCN layers are stacked. This can cause the feature of nodes to become indistinguishable, which can negatively impact classification accuracy. Additionally, shallow GCNs may struggle to effectively propagate label information throughout the entire graph when there is a limited amount of labeled nodes available \cite{li2018deeper}. Besides, GCNs require additional validation sets for hyperparameter optimization and model selection, which is incompatible with the purpose of semi-supervised learning. To address the limitations, a common approach is to use self-training to expand label set using pseudo-labels based on the confident predictions \cite{li2018deeper,sun2020multi}. This can help to improve generalization performance with limited labeled nodes and eliminate the need for an extra validation set. Nevertheless, these approachs still randomly select the initial label set, which can lead to unstable generalization performance.

\begin{figure*}[h!]
	\centering
	\begin{subfigure}{0.4\linewidth}
		\centering
		\includegraphics[width=0.7\linewidth]{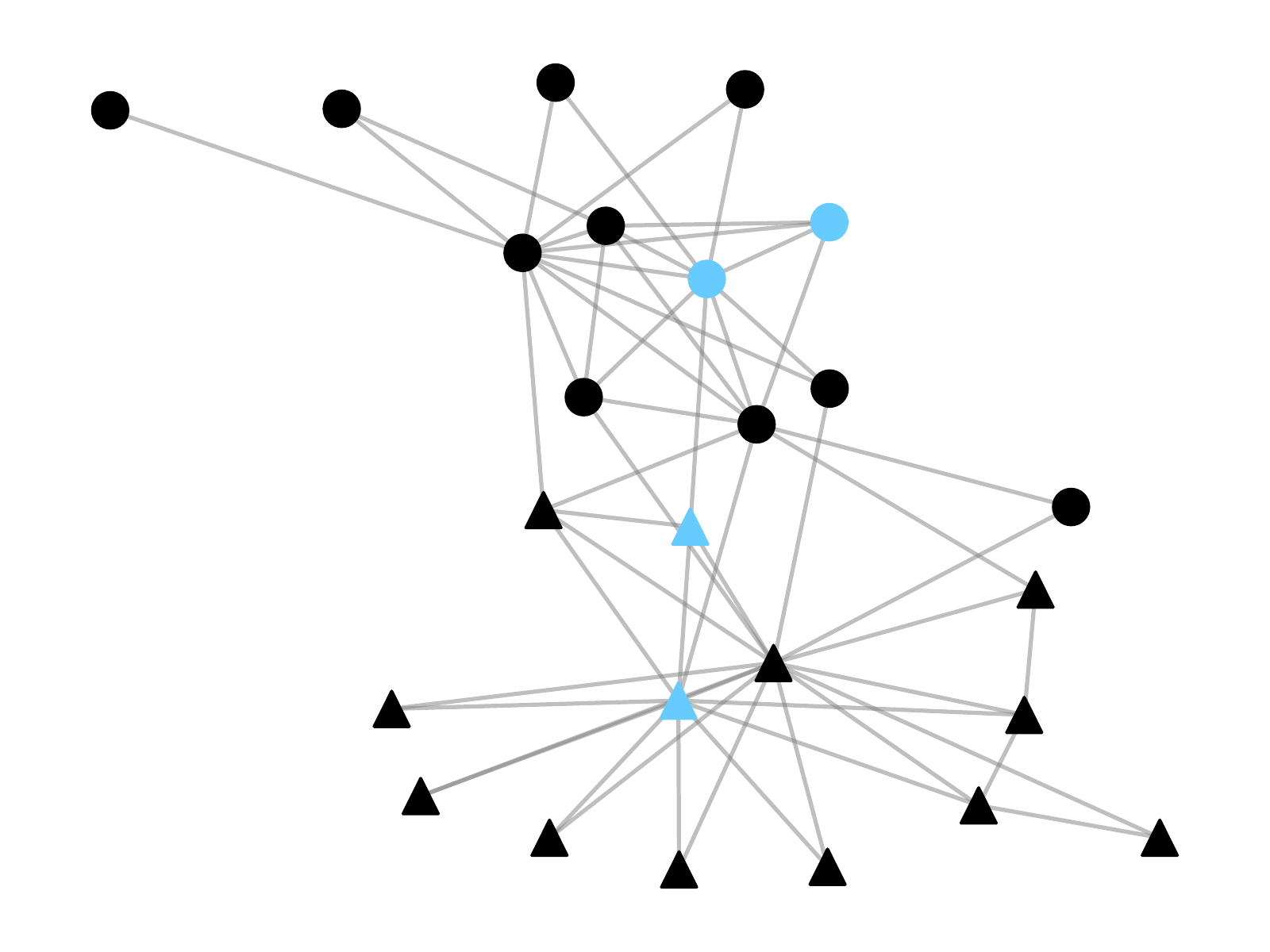}
		\caption{Node selection by confidence}
		\label{fig:Karate_confidence}
	\end{subfigure}
	\centering
	\begin{subfigure}{0.4\linewidth}
		\centering
		\includegraphics[width=0.7\linewidth]{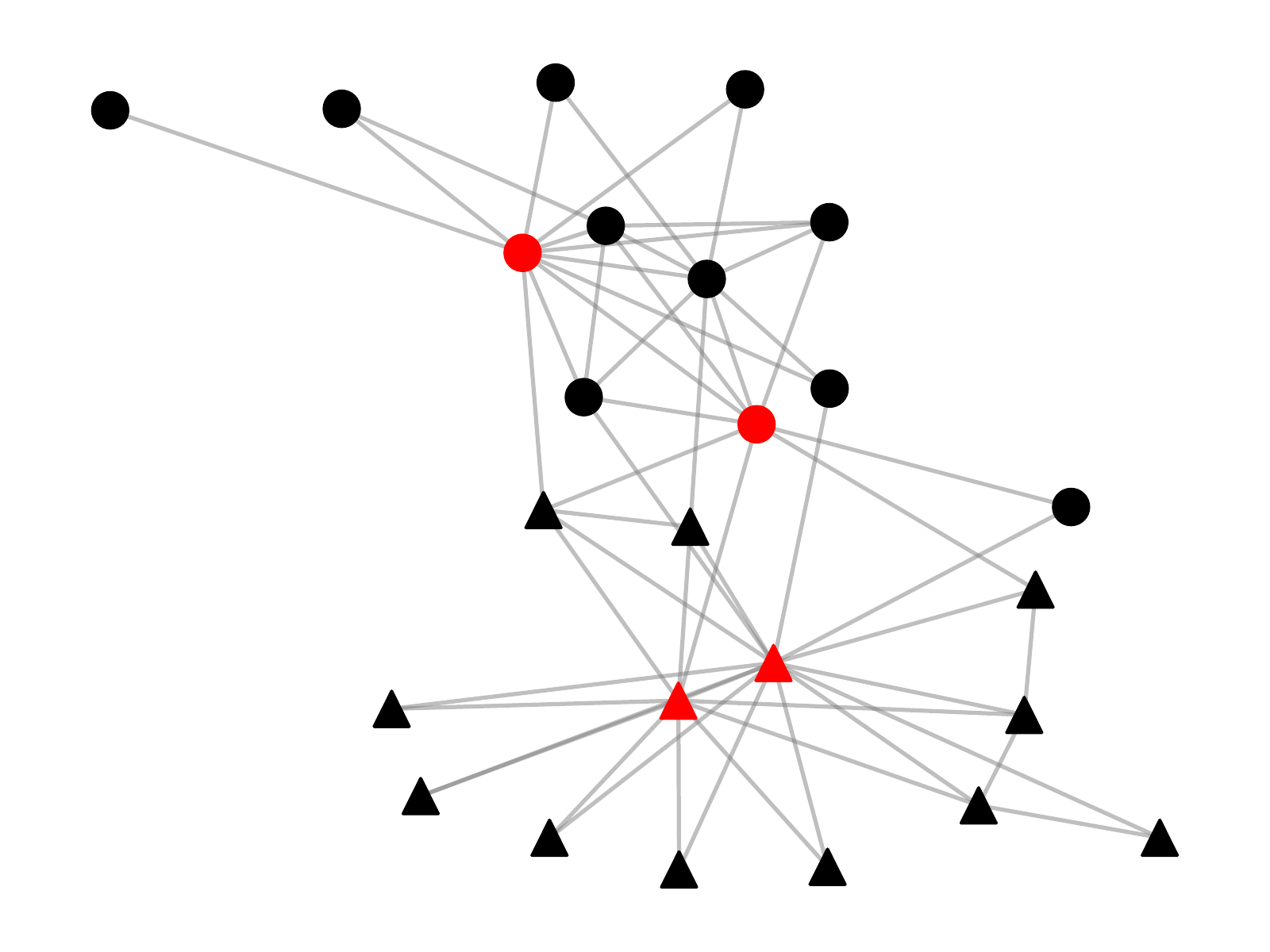}
		\caption{Node selection by DNS algorithm}
		\label{fig:Karate_DNS}
	\end{subfigure}
	\caption{Nodes selected by different methods on Zachary’s karate club. The DNS algorithm can select the typical nodes and divergent nodes based on the structural information of the graph, while the nodes selected by confidence are hardly representative of the whole subgraph.}
	\label{fig:Selection}
\end{figure*}

It should be noted that each node in a graph has the potential to represent the data distribution differently, due to the node's adjacency to other nodes in the graph. Some nodes may be more typical (i.e., have a high degree) while some nodes may be more divergent (i.e., have a low degree). To the best of our knowledge, current approaches for semi-supervised node classification using graph convolutional networks tend to simply use a random partitioning method to select the label set and the test set, without considering the typicality or divergence of the nodes in the graph. The shallow structure of GCNs leads to inefficient label information propagation, causing a decrease in the generalization ability of the model when trained with non-representative nodes. As a result, the accuracy of current methods tends to fluctuate during multiple trainings.

In this paper, we propose a graph-based deterministic node selection algorithm: Determinate Node Selection (DNS), which exploits representative nodes in the graph as label set by using the structural information of the graph. The DNS algorithm is inspired by DeLaLA \cite{xu2022semi}, which has been extensively tested and shown to be effective on Euclidean data. Unlike self-training methods that rely on pseudo-labels with high confidence, our DNS algorithm is non-iterative and determines the label set using only the node feature and structural information of the graph, without following model training. The comparison in Figure \ref{fig:Selection} illustrates the difference between the nodes selected by the DNS algorithm and those selected by the high confidence on a subgraph of two classes contained in Zachary's karate club dataset \cite{zachary1977information}. Through extensive experimentation, we demonstrate that using DNS to select the label set for GCNs leads to significant improvements in generalization performance. The main contributions of this paper are as follows:

\begin{itemize}
    \item We probe the difference in the ability of the nodes in a graph to represent the data distribution and point out that to maintain good generalization performance, it is necessary for GCNs and other semi-supervised GNNs to be trained on representative nodes.
	\item We propose a deterministic graph node selection algorithm called DNS to replace the random split method in the original GCNs. This one-shot method does not require iterative training and allows for the predetermined selection of a label set.
	\item The DNS algorithm has the versatility to be applied as a general method to arbitrary semi-supervised node classification GNNs due to its ability to determine label sets using only the node features and structural information of the graph. Our experiments demonstrate that the use of label set selected by DNS algorithm can improve model performance.
\end{itemize}

\section{Related Works}

\subsection{Graph-Based Semi-Supervised Learning}

Graph-Based Semi-Supervised Learning (GSSL) is a subfield of semi-supervised learning in which  samples in a dataset are represented as nodes in an affinity graph, and the labeling information of unlabeled samples is inferred based on the structure of the affinity graph. The anchor graph-based GSSL methods \cite{wang2016scalable,he2020fast} use the K-Means algorithm to generate anchor points on the dataset and construct an affinity graph using the similarity between data and anchor points. Optimal leading forest (OLF) \cite{xu2017local} is a structure that reveals the evolution of difference along paths within sub-trees, and the DeLaLA algorithm \cite{xu2022semi} which utilizes OLF for label propagation has achieved state-of-the-art performance on multiple datasets. However, GSSL methods are primarily designed for use with data in the Euclidean space and have not involved the study of graph data in the non-Euclidean domains.

\subsection{Convolutional Graph Neural Networks}

Convolutional Graph Neural Networks (ConvGNNs) extend the convolutional operation from grid data to graph data, generating node representations by aggregating features  from both the nodes themselves and their neighbors, and stack multiple graph convolutional layers to extract high-level node representations. ConvGNNs can be divided into two categories: Spatial-Based ConvGNNs and Spectral-Based ConvGNNs. Spatial-Based ConvGNNs \cite{hamilton2017inductive,velivckovic2018graph} define graph convolution based on the spatial relationship between nodes and propagate information between nodes through their edges. Spectral-based approaches \cite{kipf2016semi,li2018adaptive} perform graph convolution operations on the spectral domain of the graph.

GCNs \cite{kipf2016semi} have simplified the spectral decomposition and bridged the gap between spectral-based and spatial-based approaches. However, GCNs suffer from the over-smoothing, which makes GCNs unsuitable for stacking over multiple layers. Additionally, GCNs are not able to effectively propagate label information to the entire graph when there are only a few labeled nodes. These limitations may impact the performance of GCNs in certain scenarios.

\begin{figure*}[h!]
\centering 
\includegraphics[width=7in]{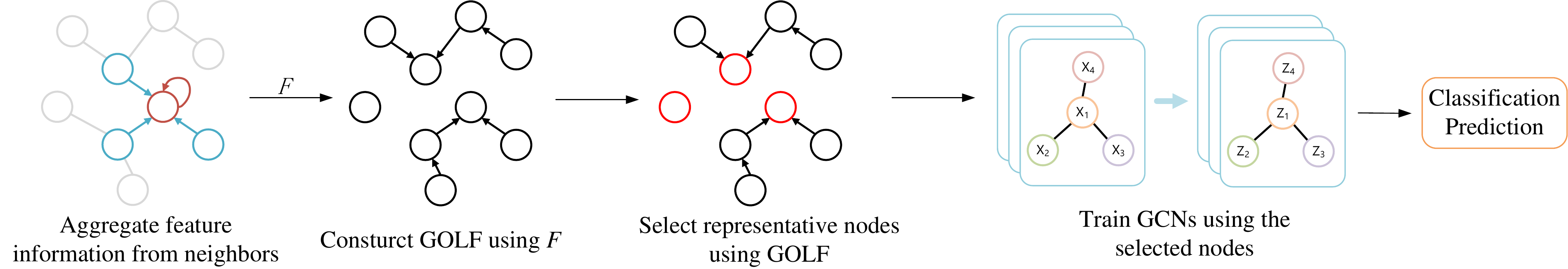}
\caption{Flowchart of DNS algorithm.}
\label{fig:Flowchart}
\end{figure*}

\subsection{Pseudo-label Methods on GCNs}

\cite{li2018deeper} identified the strengths and weaknesses of GCNs, and proposed two pseudo-label methods: Co-Training and Self-Training, as a means of expanding the label set to improve the generalization performance of GCNs in the scenario of the graph with few labeled nodes. Co-Training utilizes a random walk model to add the neighbors of labeled nodes to the label set for training GCNs. Self-Training, on the other hand, starts with the initial label set to train GCNs and then uses the classification predictions of GCNs to select the nodes with the highest confidence in each class and add them to the label set for further training. \cite{sun2020multi} proposed an improvement to the Self-Training method outlined in \cite{li2018deeper} by introducing the multi-stage training framework and the alignment mechanism to enhance the accuracy of pseudo-labels.

While the above Pseudo-label Methods have greatly improved the classification accuracy of GCNs at a low label rate, the improvement they provide over Original GCNs tends to decrease as the label rate increases. The pseudo-label Methods also have the potential for introducing noisy labels into the label set, which can negatively impact the training of GCNs. In addition, these methods rely on random split to select the initial label set, and for graphs with few labeled nodes, the quality of the obtained pseudo-labels may be compromised if the labeled nodes are not representative. This can subsequently affect the performance of the model.

\section{Determinate Node Selection on GCNs}

We found that the accuracy of GCNs can vary significantly depending on the randomly splitting label set, which motivated us to address this issue. GCNs utilize the features of unlabeled nodes in a graph for learning, but the update of gradients is reliant on labeled nodes. If the labeled nodes are not representative, the model is unable to effectively propagate labeling information to the entire graph, leading to a significant decrease in generalization performance. Inspired by DeLaLA \cite{xu2022semi}, we design a new deterministic node selection algorithm called DNS to replace the random split method and improve the stability of GCNs. The DNS algorithm allows for the classification accuracy of GCNs to converge across multiple trainings. The process of the DNS algorithm is shown in Figure \ref{fig:Flowchart}.

\subsection{Graph-based Optimal Leading Forest}

The concepts of ``multimodal" and ``mixmodal" were proposed in \cite{zhang2019semi}. In simple terms, multimodal refers to the presence of multiple clusters within a single class, while mixmodal refers to the overlap of samples between different classes within a single cluster. The randomly selected labeled nodes may only belong to one cluster within a class and not represent all nodes within that class. Additionally, due to the mixmodal nature of the data, it may be difficult to distinguish labeled nodes from nodes in other classes.

To mitigate the negative effects of mixmodal and multimodal on model training, we need to design an algorithm for selecting representative nodes from the graph. These nodes can be classified into two categories: typical nodes, which are located at the center of clusters within a class, and divergent nodes, which are found at the edges of clusters within a class. Our goal is to adaptively select a representative set of nodes that includes both typical and divergent nodes.

The Optimal leading forest (OLF) is derived from the Density Peaks Clustering Algorithm \cite{rodriguez2014clustering}. OLF is a non-iterative algorithm and is designed to represent the biased order relationship between samples by their local density, while also capturing the evolution from the central location to the edges of the samples \cite{xu2021lapoleaf}.  According to the characteristics of graph data, we developed the Graph-based Optimal Leading Forest (GOLF). GOLF uses the feature of the nodes in the graph and the structure of the graph to construct a leading tree, from which it identifies the centers of clusters and disconnects them from their parent nodes to form a leading forest. GOLF is constructed as follows.

To extract meaningful information from the graph, we apply graph convolution operations to aggregate the features of each node and its neighbors, embedding the structural information of the graph into the resulting node features. These node features serve as the input for GOLF, denoted as F:

\begin{equation}\label{eq:gcnNorm}
\begin{aligned}
F=\tilde{D}^{-\frac{1}{2}}\tilde{A}\tilde{D}^{-\frac{1}{2}}X,
\end{aligned}
\end{equation}
where $X$ denotes the node feature, $\tilde{A}=A+I$ and $\tilde{D}$ is the degree
matrix of $\tilde{A}$. $A$ is the the adjacent matrix of the graph.

$F$ contains information about the density of nodes in the graph: if a node has more neighbors, or if the feature of nearby nodes contain more information, its aggregation will be more informative. As a result, the local density of nodes in the graph can be calculated using $F$, which can be used to measure the typicality of nodes within the graph.

The local density is calculated as follows:

\begin{equation}\label{eq:density}
\begin{aligned}
\rho_i = \exp({\frac{-{\Vert F_i\Vert}_2^2}{\sigma^2}})\quad (\text{for }1\leq i\leq n),
\end{aligned}
\end{equation}
where $n$ is the size of nodes, $F_i$ is the feature of node $i$ and $\sigma$ is a bandwidth parameter.

In the leading tree, each node is connected to a leading node (the parent in the leading tree) with a higher local density, except for the node with the highest local density. We use the graph adjacency and local density to determine the leading node $Pa_i$ for each node $i$ in the leading tree:

\begin{equation}\label{eq:leadingNode}
 \resizebox{.91\linewidth}{!}{$
	\displaystyle
	Pa_i=\begin{cases} -1,& \text{if $i= \arg\max(\rho)$}\\
	\mathop{\arg\max}(\rho_j)\quad(\text{for }j\in \mathcal{N}(i)),& \text{otherwise},
	\end{cases}
	$}
\end{equation}
where $\mathcal{N}(i)$ denotes the neighbors of node $i$. The leading relationships reflect the biased order relationship between nodes in the graph. Specifically, $Pa_i$, the leading node of node $i$ is closer to the center of the graph than the node $i$. Since the input feature F contains the structural information of the graph, it makes the local density a representation of the centrality of nodes in the graph. Therefore, we use the local density to denote the distance of node $i$ from its leading node $Pa_i$:

\begin{equation}\label{eq:delta}
 \resizebox{.85\linewidth}{!}{$
	\displaystyle
	\delta_i=\begin{cases} \mathop{\min}(\rho_j) \quad(\text{for }j\in \mathcal{N}(i)),& \text{if $i= \arg\max(\rho)$}\\
	\rho_{Pa_i},& \text{otherwise}.
	\end{cases}
	$}
\end{equation}

This process results in a leading tree that includes all the nodes in the graph. To construct a GOLF, it is necessary to identify the centers of clusters in the leading tree. This centrality can be formulated as follows:

\begin{equation}\label{eq:gamma}
\gamma_i=\rho_i*\delta_i,
\end{equation}
where $\gamma_i$ denotes the probability that node $i$ will be chosen as the center of a class cluster. By disconnecting the node with the highest $\gamma$ value from its leading node, the GOLF can be constructed. This process also allows us to determine the depth $layer_i$ of each node $i$ within the GOLF.

The path of each tree in the GOLF follows the leading relationships, representing the evolution of the nodes in the graph from the edge towards the center. It is worth noting that the construction of GOLF does not require iterative optimization, making it an efficient algorithm.

\subsection{Deterministic Labeling Objective Function}

\cite{xu2022semi} proposed an efficient Deterministic Labeling method using OLF to identify the key samples in the data and to depict the underlying structure of the data as completely as possible using a small number of samples. This method utilizes the OLF structure, so that the selected set of labeled samples contains both typical and divergent samples, and has a complexity of $O(n\log n)$. We believe this method has the potential to be applied to graph data, and in this section, we provide a brief overview of the method.

The Deterministic Labeling method utilizes a discrete optimization objective function to obtain the label set. Let $\mathcal{L}$ denotes the the label set, and $\mathcal{L}_c$ denotes the labeled samples of class c. $\mathcal{L}$ can be divided into two parts: $\mathcal{L}_{Typ}$ and $\mathcal{L}_{Div}$, denote the typical and divergent samples, respectively. $\mathcal{L}$ can be obtained by solving the following objective function:

\begin{equation}\label{eq:select}
	\resizebox{.91\linewidth}{!}{$
		\begin{aligned}
		{\min}J(\mathcal{L})=&\alpha\sum_{x_{i}\in \mathcal{L}_{Typ}}q(\gamma_i)+(1-\alpha)\sum_{x_{j}\in \mathcal{L}_{Div}}				\frac{\rho_j}{layer_j},\\
		s.t.\quad &\lvert \mathcal{L} \rvert =l,\\
		&\lvert {\mathcal{L}}_c \rvert \ge k,
		\end{aligned}
	$}
\end{equation}
where the weight parameter $\alpha$ is used to adjust the emphasis on typical or divergent samples, $\vert\bullet\vert$ denotes the number of samples in the set. $l=\vert\mathcal{L}\vert$, $k$ is the minimum number of samples selected for each class, and $q(\gamma_i)=\frac{1}{\log(\gamma_i)}$.

Since $q(\gamma_i)$ is a decreasing function, it is obvious that the objective function tends to select samples with larger $\gamma$, which usually corresponds to the most typical pattern of a class or a mode within a class. Minimizing the first term in the objective function, therefore, results in the selection of samples with high typicality. Minimizing the second term in the objective function prefers samples with deeper layers and smaller local densities in the leading tree. Since the leading tree represents the biased order relationship between samples, the samples selected in this way are the most divergent samples within a class. The Deterministic Labeling method can be applied to GOLF, as it relies solely on the construction of OLF. Details of the Deterministic Labeling method can be viewed in \cite{xu2022semi} or in our code.

\begin{table*}[htb]
	\renewcommand{\thetable}{2}
    \centering
     \setlength{\tabcolsep}{0.8mm}{
    \begin{tabular}[10cm]{crrrrrr}
        \toprule
        \multicolumn{2}{c}{\textbf{Label Rate}} & 0.5\% & 1\% & 2\% & 3\% & 4\%\\ 
        \midrule
        \multicolumn{2}{c}{\textbf{LP}} & 58.0 $\pm$ 5.7 & 61.5 $\pm$ 4.8 & 63.6 $\pm$ 2.8 & 64.4 $\pm$ 2.5 & 66.1 $\pm$ 2.8 \\
        \multicolumn{2}{c}{\textbf{GCN}} & 50.6 $\pm$ 7.9 & 58.7 $\pm$ 7.8 & 70.0 $\pm$ 6.4 & 75.8 $\pm$ 3.4 & 76.6 $\pm$ 2.5 \\
        \hdashline[0.5pt/5pt]
        \multirow{4}{*}{\textbf{DI}}
        & \textbf{Co-training} & 53.9 $\pm$ 5.4 & 59.9 $\pm$ 3.7 & 71.5 $\pm$ 3.3 & 75.1 $\pm$ 2.8 & 75.7 $\pm$ 2.3 \\
        & \textbf{Self-training} & 57.1 $\pm$ 7.6 & 63.1 $\pm$ 8.7 & 71.7 $\pm$ 4.4 & 76.8 $\pm$ 1.9 & 77.5 $\pm$ 1.2 \\
        & \textbf{Union} & 55.5 $\pm$ 5.4 & 62.8 $\pm$ 9.2 & 72.2 $\pm$ 3.4 & 76.9 $\pm$ 2.8 & 77.7 $\pm$ 1.6 \\
        & \textbf{Intersection} & 50.2 $\pm$ 8.3 & 63.2 $\pm$ 6.0 & 70.6 $\pm$ 4.4 & 74.7 $\pm$ 2.4 & 76.0 $\pm$ 2.0 \\
        \hdashline[0.5pt/5pt]
        \multicolumn{2}{c}{\textbf{M3S}} & 61.6 $\pm$ 7.1 & 67.7 $\pm$ 5.6 & 75.8 $\pm$ 3.5 & 77.8 $\pm$ 1.6 & 78.0 $\pm$ 2.1 \\
        \midrule
        \multicolumn{2}{c}{\textbf{LP+DNS}} & (+5.3) $\pm$ (-4.5) & (+4.1) $\pm$ (-3.5) & (+1.1) $\pm$ (-1.7) & (+0.6) $\pm$ (-1.0) & (+4.0) $\pm$ (-1.9) \\
        \multicolumn{2}{c}{\textbf{GCN+DNS}} & (+10.0) $\pm$ (-2.6) & (+8.3) $\pm$ (-3.8) & (+6.2) $\pm$ (-4.3) & (+3.4) $\pm$ (-2.7) & (+4.5) $\pm$ (-1.0) \\
        \hdashline[0.5pt/5pt]
        \multirow{4}{*}{\textbf{DI+DNS}}
        & \textbf{Co-training} & (+0.8) $\pm$ (-1.6) & (+5.9) $\pm$ (-1.8) & (+1.7) $\pm$ (-3.9) & (+1.7) $\pm$ (-1.2) & (+4.9) $\pm$ (-1.7) \\
        & \textbf{Self-training} & (+0.6) $\pm$ (-4.1) & (+2.3) $\pm$ (-6.3) & (+1.7) $\pm$ (-1.8) & (+0.8) $\pm$ (-0.6) & (+2.1) $\pm$ (-0.3) \\
        & \textbf{Union} & (+2.3) $\pm$ (-1.4) & (+1.7) $\pm$ (-6.6) & (+2.4) $\pm$ (-1.8) & (+1.0) $\pm$ (-2.0) & (+3.7) $\pm$ (-0.7) \\
        & \textbf{Intersection} & (+1.7) $\pm$ (-4.0) & (+2.6) $\pm$ (-3.9) & (+1.1) $\pm$ (-2.3) & (+0.7) $\pm$ (-0.4) & (+3.2) $\pm$ (-1.2) \\
        \hdashline[0.5pt/5pt]
        \multicolumn{2}{c}{\textbf{M3S+DNS}} & (+2.7) $\pm$ (-3.4) & (+1.6) $\pm$ (-2.5) & (+0.6) $\pm$ (-0.7) & (+0.4) $\pm$ (-0.4) & (+0.6) $\pm$ (-0.8) \\
        \bottomrule
    \end{tabular}}
	\caption{Mean classification accuracy and standard deviation (\%) on Cora with a few labeled nodes.}
	\label{tab:low_labelrate_Cora}
	\vspace{-7pt} 
\end{table*}

\begin{table*}[htb]
	\renewcommand{\thetable}{3}
    \centering
     \setlength{\tabcolsep}{0.8mm}{
    \begin{tabular}[10cm]{crrrrrr}
        \toprule
        \multicolumn{2}{c}{\textbf{Label Rate}} & 0.5\% & 1\% & 2\% & 3\% & 4\%\\ 
        \midrule
        \multicolumn{2}{c}{\textbf{LP}} & 37.8 $\pm$ 7.5 & 41.8 $\pm$ 4.5 & 42.3 $\pm$ 4.7 & 44.7 $\pm$ 2.6 & 44.8 $\pm$ 2.4 \\
        \multicolumn{2}{c}{\textbf{GCN}} & 45.1 $\pm$ 8.4 & 54.7 $\pm$ 4.6 & 61.6 $\pm$ 4.3 & 67.1 $\pm$ 2.8 & 69.0 $\pm$ 2.7 \\
        \hdashline[0.5pt/5pt]
        \multirow{4}{*}{\textbf{DI}}
        & \textbf{Co-training} & 44.1 $\pm$ 9.8 & 51.1 $\pm$ 8.8 & 58.9 $\pm$ 7.0 & 64.7 $\pm$ 2.7 & 65.6 $\pm$ 2.0 \\
        & \textbf{Self-training} & 51.6 $\pm$ 6.3 & 57.7 $\pm$ 4.6 & 64.8 $\pm$ 2.2 & 67.9 $\pm$ 1.6 & 68.8 $\pm$ 2.1 \\
        & \textbf{Union} & 48.7 $\pm$ 5.8 & 55.1 $\pm$ 4.1 & 61.8 $\pm$ 5.3 & 66.6 $\pm$ 2.2 & 67.1 $\pm$ 2.2 \\
        & \textbf{Intersection} & 51.8 $\pm$ 8.7 & 61.1 $\pm$ 4.5 & 63.0 $\pm$ 2.7 & 69.5 $\pm$ 1.3 & 70.1 $\pm$ 2.0 \\
        \hdashline[0.5pt/5pt]
        \multicolumn{2}{c}{\textbf{M3S}} & 56.2 $\pm$ 5.6 & 62.4 $\pm$ 4.6 & 66.5 $\pm$ 2.5 & 70.3 $\pm$ 2.0 & 70.6 $\pm$ 1.9 \\
        \midrule
        \multicolumn{2}{c}{\textbf{LP+DNS}} & (+4.1) $\pm$ (-5.9) & (+3.0) $\pm$ (-3.5) & (+1.5) $\pm$ (-3.4) & (+0.5) $\pm$ (-1.1) & (+1.0) $\pm$ (-1.0) \\
        \multicolumn{2}{c}{\textbf{GCN+DNS}} & (+7.0) $\pm$ (-4.4) & (+7.3) $\pm$ (-2.3) & (+2.5) $\pm$ (-2.8) & (+0.9) $\pm$ (-1.8) & (+1.2) $\pm$ (-1.2) \\
        \hdashline[0.5pt/5pt]
        \multirow{4}{*}{\textbf{DI+DNS}}
        & \textbf{Co-training} & (+5.6) $\pm$ (-7.5) & (+5.4) $\pm$ (-6.9) & (+0.5) $\pm$ (-5.5) & (+1.6) $\pm$ (-1.8) & (+1.5) $\pm$ (-1.0) \\
        & \textbf{Self-training} & (+4.2) $\pm$ (-2.4) & (+1.0) $\pm$ (-1.6) & (+0.6) $\pm$ (-0.2) & (+0.6) $\pm$ (-0.2) & (+1.4) $\pm$ (-1.1) \\
        & \textbf{Union} & (+2.2) $\pm$ (-2.1) & (+3.1) $\pm$ (-2.2) & (+2.2) $\pm$ (-2.5) & (+2.3) $\pm$ (-0.9) & (+1.0) $\pm$ (-0.7) \\
        & \textbf{Intersection} & (+7.6) $\pm$ (-4.9) & (+1.1) $\pm$ (-2.5) & (+2.9) $\pm$ (-0.8) & (+0.8) $\pm$ (-0.2) & (+0.4) $\pm$ (-0.7) \\
        \hdashline[0.5pt/5pt]
        \multicolumn{2}{c}{\textbf{M3S+DNS}} & (+5.7) $\pm$ (-3.5) & (+1.1) $\pm$ (-1.4) & (+1.1) $\pm$ (-0.7) & (+0.4) $\pm$ (-0.6) & (+0.2) $\pm$ (-0.4) \\
        \bottomrule
    \end{tabular}}
	\caption{Mean classification accuracy and standard deviation (\%) on Citeseer with a few labeled nodes.}
	\label{tab:low_labelrate_Citeseer}
	\vspace{-7pt} 
\end{table*}

\IncMargin{1em} 
\begin{algorithm}[!htb]
    \caption{Determinate Node Selection Algorithm}
    \label{alg:DNS}
    \begin{spacing}{1.3}
    \KwIn{Node features matrix $X$, adjacent matrix $A$,node set $N$, size of labeled node set $l$}
    \KwOut{labeled node set $L$}
    \tcp{Step 1: Compute input feature}
	$F=\tilde{D}^{-\frac{1}{2}}\tilde{A}\tilde{D}^{-\frac{1}{2}}X$\\
	\tcp{Step 2: Consturct GOLF}
    $\rho = e^{\frac{-{\Vert F\Vert}_2^2}{\sigma^2}}$\\
	\For{node $i$ in $N$}
	{    \resizebox{.89\linewidth}{!}{$
		 \displaystyle
		 Pa_i=\begin{cases} -1,& \text{if $i= \arg\max(\rho)$}\\
		\mathop{\arg\max}(\rho_j)\quad(\text{for }j\in \mathcal{N}(i)),& \text{otherwise},
		\end{cases}
		$}
		 \resizebox{.85\linewidth}{!}{$
		\displaystyle
		\delta_i=\begin{cases} \mathop{\min}(\rho_j) \quad(\text{for }j\in \mathcal{N}(i)),& \text{if $i= \arg\max(\rho)$}\\
	\rho_{Pa_i},& \text{otherwise}.
	\end{cases}
		$}
	}
	$\gamma_i=\rho_i*\delta_i$\\
	Compute $layer_i$ using $\gamma$ to consturct GOLF.\\
    \tcp{Step 3: Select $l$ nodes to label}
    $L=Deterministic\_Labeling(\gamma,\rho,layer_i)$\\
    \textbf{return} $L$
    \end{spacing}
	\vspace{-7pt} 
\end{algorithm}

\subsection{Determinate Node Selection Algorithm}

In summary, the DNS algorithm is described in Algorithm \ref{alg:DNS}. The label set obtained using the DNS algorithm can be used as an alternative to randomly selected labels for any semi-supervised nodal classification GNNs, without requiring modifications to the original model. In the Experiments section, we report the improvement of DNS algorithm on the generalization performance of the original model.

\subsection{Complexity Analysis}

The DNS algorithm consists of three main parts: computing the input features $F$, constructing the GOLF, and selecting the labeled nodes. For a graph with $e$ edges, $n$ nodes, and $d$ dimensional node features, the computational complexity of computing the input features F is $O(ed)$. The complexity of constructing the GOLF is $O(ne)$, which depends only on the number of nodes and edges in the graph. The process of selecting the labeled nodes involves sorting based on the $\gamma$ of the nodes, with a complexity of $O(n\log n)$. It can be seen that the complexity of the DNS algorithm is primarily determined by the size of the graph, specifically the number of nodes and edges.

\begin{table*}[htbp]
	\renewcommand{\thetable}{4}
    \centering
     \setlength{\tabcolsep}{2.0mm}{
    \begin{tabular}[10cm]{crrrr}
        \toprule
        \multicolumn{2}{c}{\textbf{Label Rate}} & 0.03\% & 0.05\% & 0.1\% \\ 
        \midrule
        \multicolumn{2}{c}{\textbf{LP}} & 58.8 $\pm$ 9.4 & 61.5 $\pm$ 8.1 & 64.1 $\pm$ 5.2 \\
        \multicolumn{2}{c}{\textbf{GCN}} & 51.6 $\pm$ 7.5 & 58.0 $\pm$ 5.6 & 67.7 $\pm$ 6.2 \\
        \hdashline[0.5pt/5pt]
        \multirow{4}{*}{\textbf{DI}}
        & \textbf{Co-training} & 56.2 $\pm$ 10.9 & 61.6 $\pm$ 5.7 & 68.0 $\pm$ 6.4 \\
        & \textbf{Self-training} & 57.1$\pm$ 11.8 & 63.9 $\pm$ 7.3 & 70.1 $\pm$ 5.3 \\
        & \textbf{Union} & 57.3 $\pm$ 9.8 & 64.5 $\pm$ 6.1 & 70.0 $\pm$ 5.6 \\
        & \textbf{Intersection} & 55.1 $\pm$ 9.4 & 58.9 $\pm$ 7.3 & 67.9 $\pm$ 6.5 \\
        \hdashline[0.5pt/5pt]
        \multicolumn{2}{c}{\textbf{M3S}} & 60.0 $\pm$ 8.3 & 64.4 $\pm$ 10.9 & 70.6 $\pm$ 4.1 \\
        \midrule
        \multicolumn{2}{c}{\textbf{LP+DNS}} & (+4.9) $\pm$ (-8.0) & (+5.0) $\pm$ (-6.7) & (+3.6) $\pm$ (-3.5) \\
        \multicolumn{2}{c}{\textbf{GCN+DNS}} & (+5.1) $\pm$ (-5.0) & (+11.5) $\pm$ (-3.6) & (+6.6) $\pm$ (-4.0) \\
        \hdashline[0.5pt/5pt]
        \multirow{4}{*}{\textbf{DI+DNS}}
        & \textbf{Co-training} & (+5.6) $\pm$ (-7.5) & (+5.4) $\pm$ (-3.7) & (+5.8) $\pm$ (-3.5) \\
        & \textbf{Self-training} & (+6.6) $\pm$ (-6.7) & (+6.5) $\pm$ (-1.7) & (+4.0) $\pm$ (-3.2) \\
        & \textbf{Union} & (+10.2) $\pm$ (-4.5) & (+7.5) $\pm$ (-1.0) & (+5.7) $\pm$ (-3.9) \\
        & \textbf{Intersection} & (+4.0) $\pm$ (-3.9) & (+1.1) $\pm$ (-3.3) & (+3.3) $\pm$ (-2.5) \\
        \hdashline[0.5pt/5pt]
        \multicolumn{2}{c}{\textbf{M3S+DNS}} & (+8.4) $\pm$ (-3.2) & (+4.8) $\pm$ (-6.8) & (+1.0) $\pm$ (-2.6) \\
        \bottomrule
    \end{tabular}}
	\caption{Mean classification accuracy and standard deviation (\%) on Pubmed with a few labeled nodes.}
	\label{tab:low_labelrate_Pubmed}
\end{table*}

\begin{table}[h]
	\renewcommand{\thetable}{1}
	\centering
    \begin{tabular}{lrrrrr}
        \toprule
        \textbf{Dataset} & \textbf{Nodes} & \textbf{Edges} & \textbf{Classes} & \textbf{Features}\\ 
        \midrule
        Cora & 2708 & 5429 & 7 & 1433 \\
        Citeseer & 3327 & 4732 & 6 & 3703 \\
        Pubmed & 19717 & 44338 & 3 & 500 \\
        \bottomrule
    \end{tabular}
	\caption{Dateset statistics}
	\label{tab:dataset}
	\vspace{-7pt} 
\end{table}

\section{Experiments}

To confirm the effectiveness of our theoretical and DNS algorithms, we conducted experiments on three widely-used graph datasets: Cora, Citesser, and Pubmed \cite{sen2008collective}. The statistics of these datasets are summarized in Table \ref{tab:dataset}.

As for baseline, we compare label propagation using ParWalks (LP) \cite{wu2012learning}; Convolutional Networks (GCNs) \cite{kipf2016semi}; And two methods based on pseudo-label: DI \cite{li2018deeper} (including Self-Training, Co-Training, Union and Intersection) and M3S \cite{sun2020multi}.

\subsection{Experimental Setup}

\begin{figure*}[h!]
	\centering
	\begin{subfigure}{0.325\linewidth}
		\centering
		\includegraphics[width=0.9\linewidth]{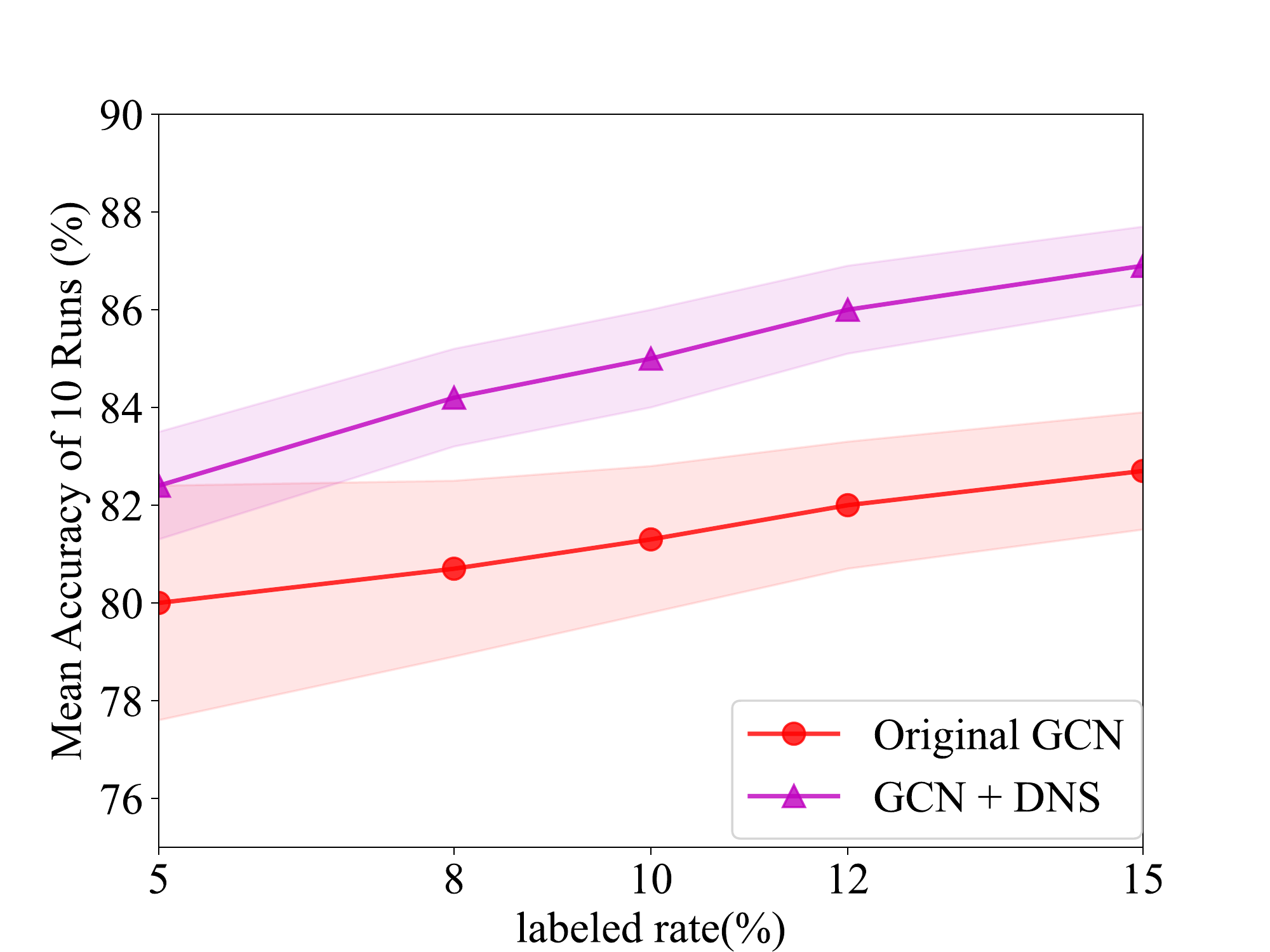}
		\caption{Cora}
		\label{fig:GCN_Cora_HighLabelRate}
	\end{subfigure}
	\centering
	\begin{subfigure}{0.325\linewidth}
		\centering
		\includegraphics[width=0.9\linewidth]{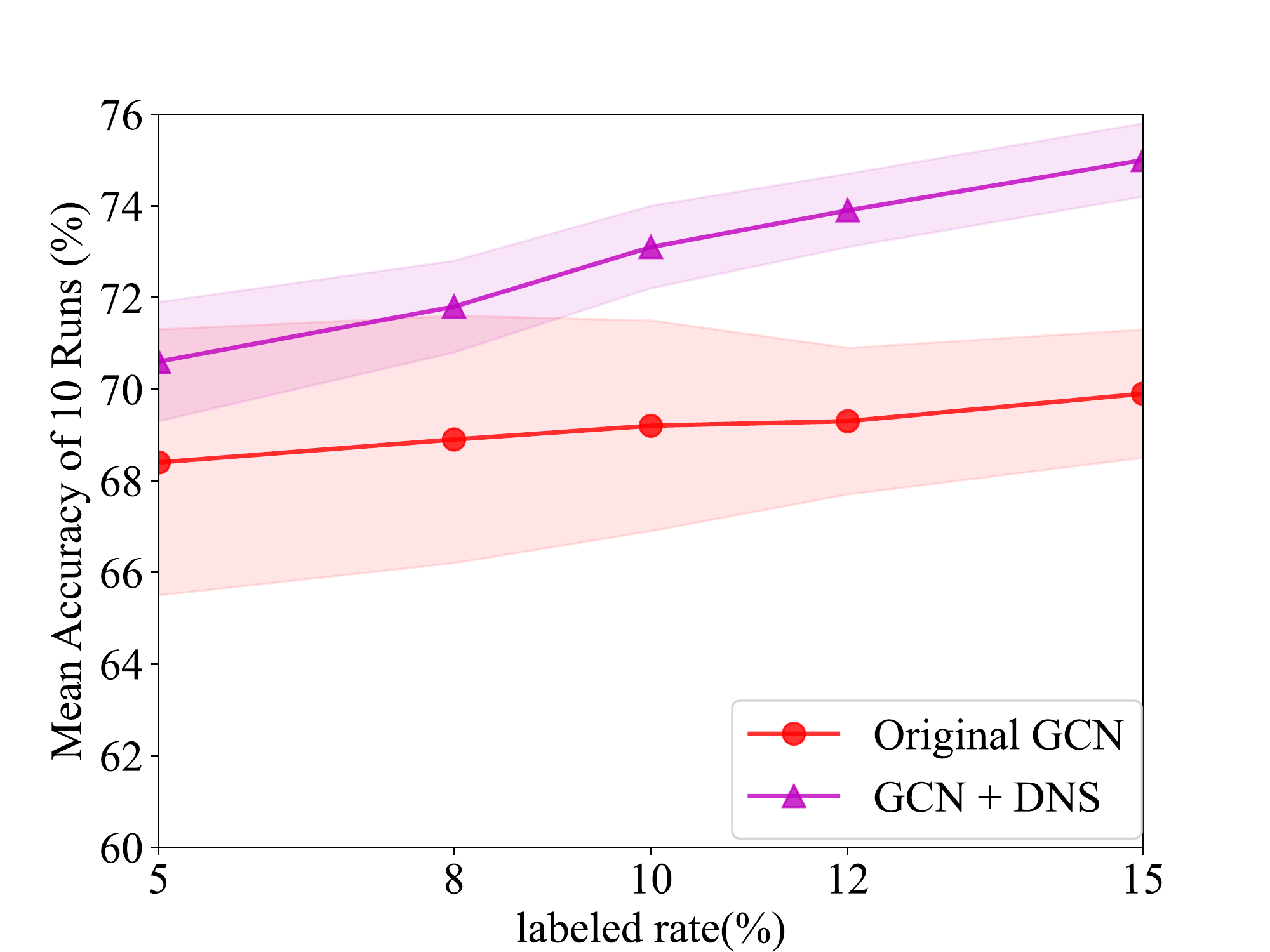}
		\caption{Citeseer}
		\label{fig:GCN_Citeseer_HighLabelRate}
	\end{subfigure}
	\centering
	\begin{subfigure}{0.325\linewidth}
		\centering
		\includegraphics[width=0.9\linewidth]{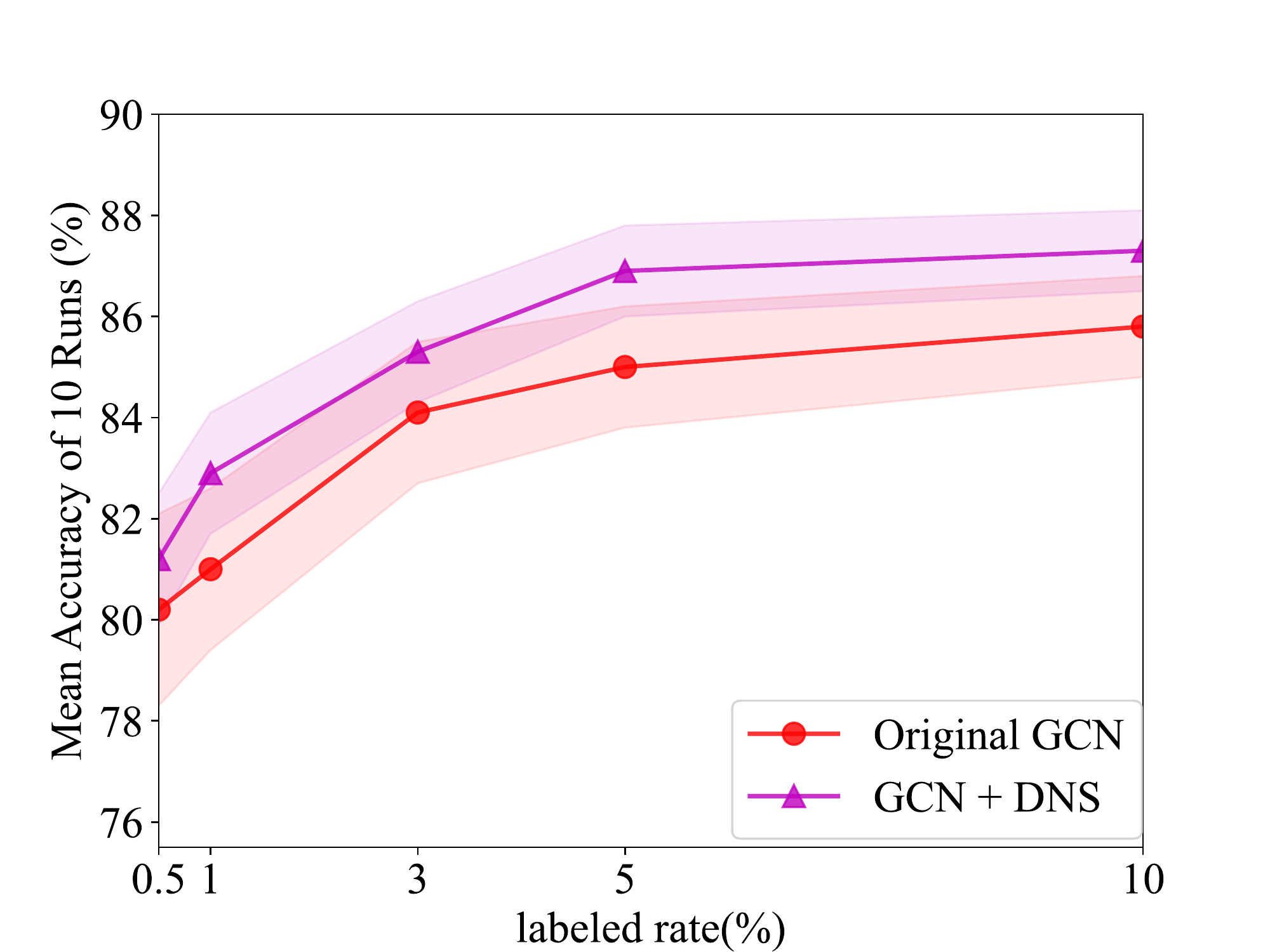}
		\caption{Pubmed}
		\label{fig:GCN_Pubmed_HighLabelRate}
	\end{subfigure}
	\caption{Mean classification accuracy on high label rate.}
	\label{fig:highLabelRate}
\end{figure*}

For LP, we employ the same hyper-parameters as in \cite{li2018deeper}. For GCNs, DI and M3S, we use the same hyper-parameters as in \cite{kipf2016semi} and \cite{sun2020multi} on Cora, Citeseer, and Pubmed: a learning rate of 0.01, 200 training epochs, 0.5 dropout rate, $5\times10^{-4}$ $L_2$ regularization weight, and 16 hidden units. The number of layers on Cora and Citeseer is 4,3,3,2,2 and 3,3,3,2,2 respectively for 0.5\%,1\%,2\%,3\%,4\% label rates, fixed 4 for 0.03\%,0.05\%,0.1\% label rates on Pubmed. We do not use an additional validation set. The K Stage setting of M3S is the same as in \cite{sun2020multi} on Cora, Citeseer, and Pubmed.

All results are the mean accuracy and standard deviation of 10 runs. For the original models, we randomly split labels into a small set for training, and a set with 1000 nodes for testing in each run. For models with DNS Algorithm, we use the same label set derived from DNS Algorithm for training and randomly select a set with 1000 nodes for testing in each run. We use the same random seed for original models and models with DNS Algorithm for fair comparison.

\begin{figure*}[htbp]
	\centering
	\begin{subfigure}{0.24\linewidth}
		\centering
		\includegraphics[width=1.12\linewidth]{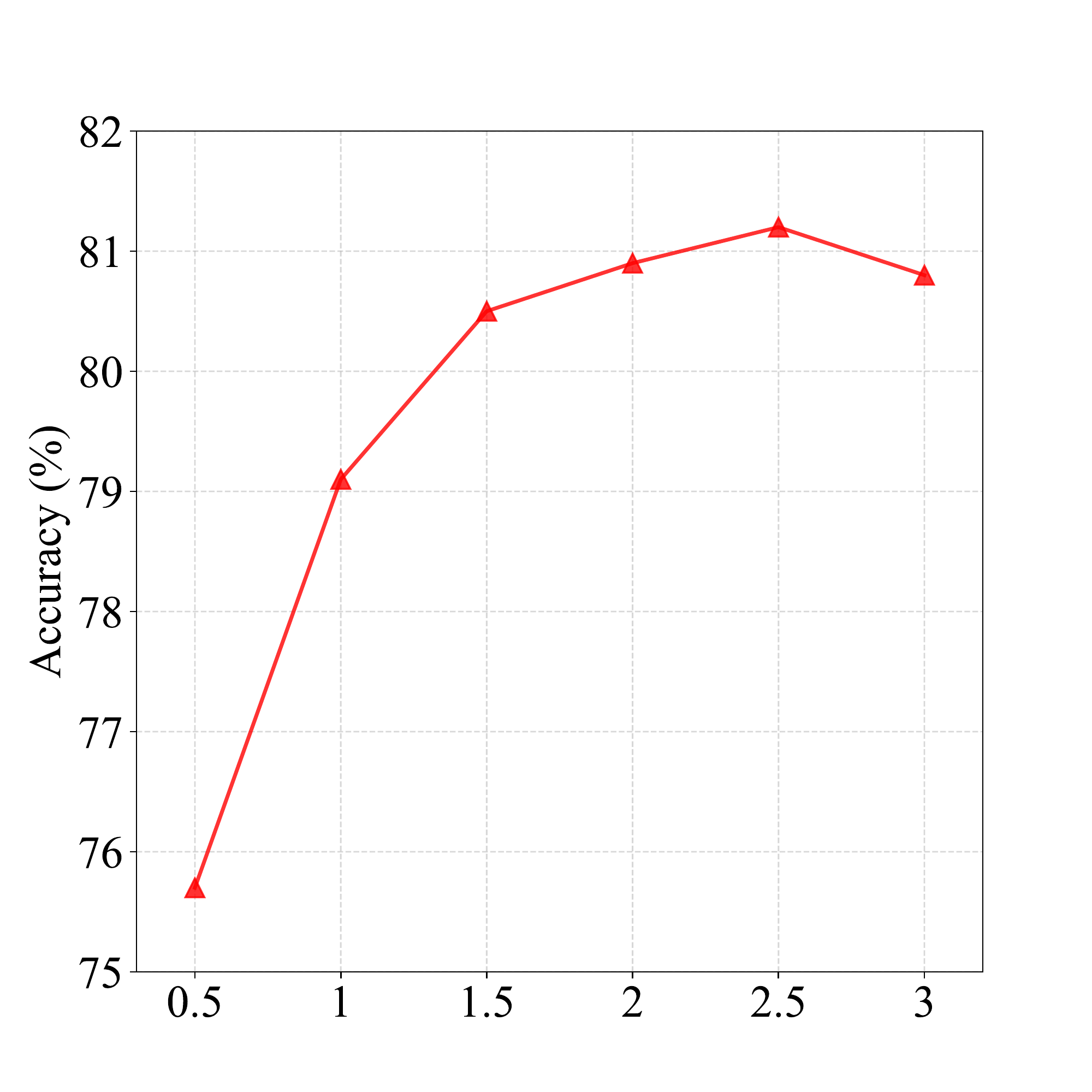}
		\caption{$\rho$}
		\label{fig:GCN_Cora_rho}
	\end{subfigure}
	\centering
	\begin{subfigure}{0.24\linewidth}
		\centering
		\includegraphics[width=1.12\linewidth]{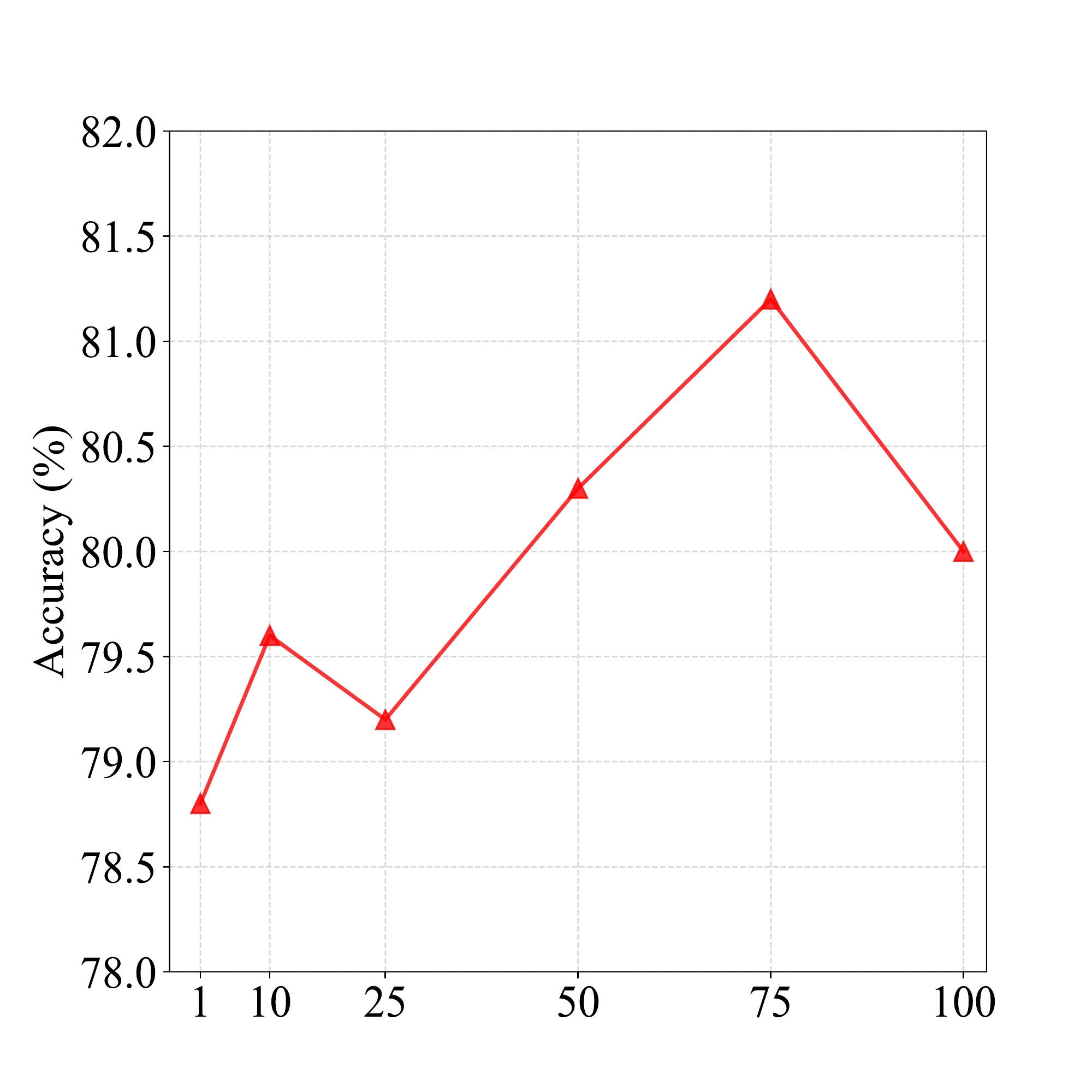}
		\caption{$n$}
		\label{fig:GCN_Cora_ltnum}
	\end{subfigure}
	\centering
	\begin{subfigure}{0.24\linewidth}
		\centering
		\includegraphics[width=1.12\linewidth]{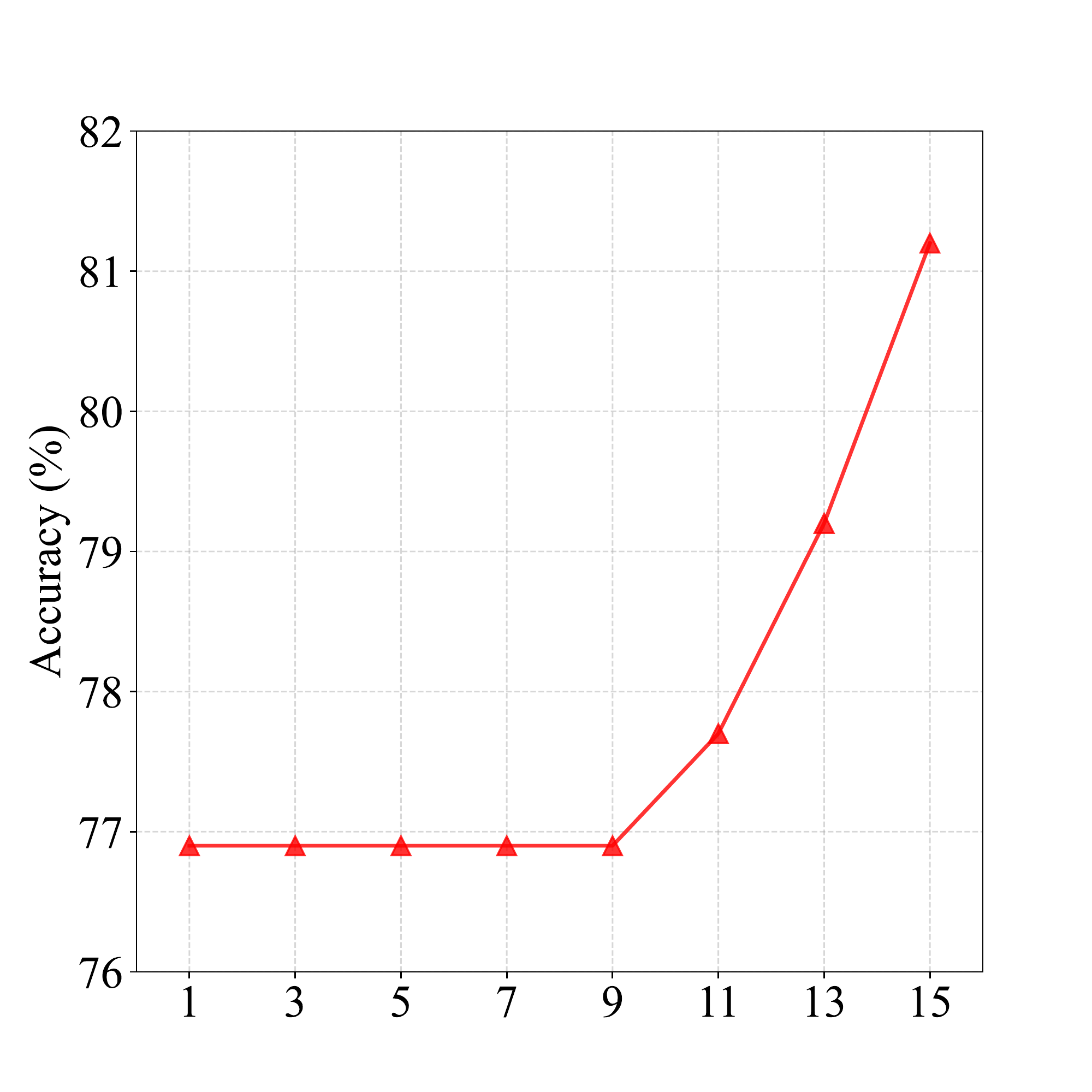}
		\caption{$k$}
		\label{fig:GCN_Cora_k}
	\end{subfigure}
	\centering
	\begin{subfigure}{0.24\linewidth}
		\centering
		\includegraphics[width=1.12\linewidth]{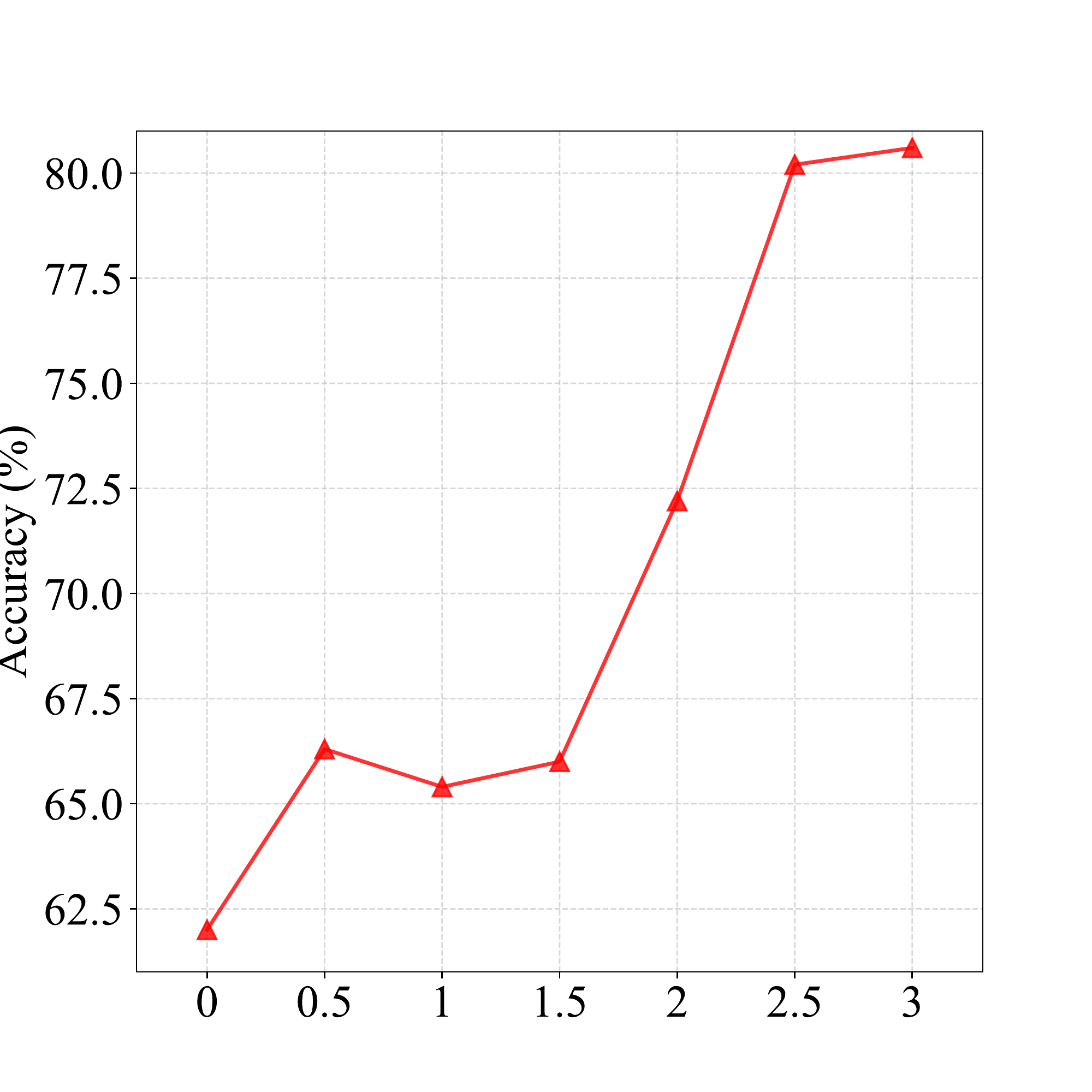}
		\caption{$\alpha$}
		\label{fig:GCN_Cora_rootweight}
	\end{subfigure}
	\caption{Parameter sensitivity.}
	\label{fig:Parameter}
\end{figure*}

\subsection{Performance of DNS Algorithm}

We report the performance improvement relative to the original method after using DNS Algorithm. The experimental results are shown in Tables \ref{tab:low_labelrate_Cora}, \ref{tab:low_labelrate_Citeseer} and \ref{tab:low_labelrate_Pubmed}, with our method displayed in the lower portion of each table.

The results demonstrate that the classification accuracy of all methods improves and the standard deviation converges to a stable interval when using the DNS Algorithm instead of a randomly divided label set. This is because the DNS Algorithm considers the typicality and divergence of nodes in the graph, and selects the most representative nodes as the label set. In addition, the lower the label rate, the more obvious the performance improvement of the DNS Algorithm.  This highlights the importance of selecting representative nodes for training and demonstrates that the DNS Algorithm is effective at identifying such nodes through exploration of the graph structure.

For pseudo-label methods DI and M3S, the gap between them and the original GCNs decreases as the labeling rate increases. We conducted experiments on the original GCNs and GCNs with the DNS Algorithm on Cora, Citeseer, and Pubmed at higher label rates, with the number of layers fixed at 2 and other parameters held constant. The results are shown in Figure \ref{fig:highLabelRate}. It can be seen that even at higher label rates, the use of the DNS Algorithm to select the label set still leads to an improvement in the generalization performance of the GCNs.

In Section 3.4, we analyzed the complexity of the DNS Algorithm. The time cost of the DNS Algorithm is less than the GCN training time for all four graph datasets. This demonstrates that our method is efficient.

\subsection{Parameter Sensitivity Analysis}

We performed sensitivity analysis on four main parameters in the DNS algorithm, including the local density $\rho$ for constructing GOLF, the number of trees in the forest $n$, the number of typical nodes $k$ selected per class, and the parameter $\alpha$ for balancing the selection of typical and divergent nodes. We conducted experiments on the Cora dataset at a label rate of 4\% and the results are shown in Figure \ref{fig:Parameter}. All four parameters affect the classification accuracy, with the DNS algorithm being particularly sensitive to $k$ and $\alpha$. Using incorrect values for $k$ and $\alpha$ can significantly decrease the classification accuracy, while for $\rho$ and $n$, the classification accuracy obtained for any value is generally higher than that of the original GCN. The experiments indicate that the DNS algorithm is a parameter-sensitive algorithm, and its parameters must be carefully tuned, especially $k$ and $\alpha$, to achieve optimal performance.

\section{Conclusion}

This paper presents the first theoretical analysis of label set selection for semi-supervised node classification GNNs. We identify the limitations of randomly splitting the label set and highlight the impact of node selection on the generalization performance of GCNs. To address these issues, we propose the DNS algorithm, which selects representative nodes in the graph by considering their typicality and divergence. We demonstrate the effectiveness of the DNS algorithm through experiments. The DNS algorithm leads to an improvement in the generalization performance of the model at various label rates. As it only uses the graph structure for node selection and is not dependent on the specific model, it can be used as a generalized preprocessing method for other semi-supervised node classification GNNs. DNS algorithm is an efficient algorithm, but it still suffers from parameter sensitivity and requires careful tuning of parameters to obtain good performance. However, the DNS algorithm exhibits parameter sensitivity and requires careful parameter tuning to achieve optimal performance. In the future, we will investigate the possibility of applying the DNS algorithm to more semi-supervised node classification GNNs, and aim to design new algorithms to automatically adjust the parameters in the DNS algorithm.

\appendix

\section*{Acknowledgments}

This work has been supported by the National Key Research and Development Program of China under grant XXX, the National Natural Science Foundation of China under grants XXX and XXX.

\bibliographystyle{named}
\bibliography{DNS}

\end{document}